\documentclass{article}
\usepackage{absspconf,amsmath,epsfig}
\usepackage{soul,color}
\usepackage{graphicx}
\usepackage{gensymb}

\DeclareUnicodeCharacter{2009}{\,}

\title{Autoencoder-based Radio Frequency Interference mitigation for SMAP Passive Radiometer}

\begin{document}
%
{
\maketitle
}
\begin{abstract}

Passive space-borne radiometers operating in the 1400-1427 MHz protected frequency band face radio frequency interference (RFI) from terrestrial sources. With the growth of wireless devices and the appearance of new technologies, the possibility of sharing this spectrum with other technologies would introduce more RFI to these radiometers. This band could be an ideal mid-band frequency for 5G and Beyond, as it offers high capacity and good coverage. Current RFI detection and mitigation techniques at SMAP (Soil Moisture Active Passive) depend on correctly detecting and discarding or filtering the contaminated data leading to the loss of valuable information, especially in severe RFI cases. In this paper, we propose an autoencoder-based RFI mitigation method to remove the dominant RFI caused by potential coexistent terrestrial users (i.e., 5G base station) from the received contaminated signal at the passive receiver side, potentially preserving valuable information and preventing the contaminated data from being discarded.  

\end{abstract}
\section{Introduction}
\label{sec:intro}

The 27MHz-wide protected portion of the L-band at 1.413 GHz has been originally set aside for radio astronomy, space research, and geophysical remote sensing. Space-borne radiometry missions such as Soil Moisture Active Passive (SMAP)  \cite{5460980} and Soil Moisture and Ocean Salinity (SMOS) \cite{kerr2001soil}, which operate in this narrow band, obtain data regarding soil moisture, ocean and surface salinity that are critical for a better understanding of the earth's hydrological cycle. Such measurements are practical because the surface emissivity at L-band is sensitive to soil moisture \cite{5460980}.

The objective of the SMAP mission is to measure the earth's soil moisture and freeze/thaw states using an active radar and a passive radiometer. SMAP scans the earth with a footprint of 40 square km to measure thermal radiation from the earth. As the earth's thermal radiation depends on soil moisture, SMAP can acquire soil moisture by measuring the intensity of thermal radiation from an area. However, radio frequency interference (RFI) can significantly compromise SMAP's radiometer measurements. Although the 1.4-1.427 GHz band is strictly protected, RFI still exists in this frequency band due to violations of International Telecommunication Union regulations and frequency leakage from adjacent bands.

Furthermore, with the ever-growing number of wireless devices, emerging bandwidth-hungry communication technologies, and the limited radio spectrum, much research and discussions have been made to decrease the crowdedness of the limited spectrum, particularly on sub-6GHz bands. One proposed solution is to release some restrictions on underutilized or fully protected radio bands by developing efficient coexistence strategies that enable the operation of multiple communication technologies on the same spectrum bands simultaneously without negatively impacting one another. While many spectrum-sharing strategies between different active communication technologies have been proposed, such methods based on spectrum sensing or databases often do not apply to the coexistence of active and passive users. Developing innovative technologies to allow spectrum sharing among active and passive users has recently become the center of several research studies \cite{nature}. Terrestrial radio telescopes can be protected from active transmitters' harmful interference if the transmitters are at a large enough distance. However, to protect orbiting remote sensing satellites currently operating on exclusive bands from RFI when/if co-channel or adjacent channel use is allowed, reliable dynamic spectrum sharing at active transmitters and efficient RFI mitigation techniques at passive radiometers are required.
The 1400-1427 MHz band is of interest for active communication for several reasons, including rain resiliency, requiring smaller antennas, and lower-cost radio equipment \cite{9983493,Piepmeier}. While authorizing active users on this band can impact the highly sensitive radiometry equipment, such coexistence can be foreseen in the future considering the fact that terrestrial telescopes can be guarded against RFI by separating the active transmitters from them, or the remote orbiting satellites which are present above a particular area for only limited time can be potentially protected through opportunistic access strategies or effective power control mechanisms at active receivers.

Current SMAP measurements show significant RFI, while this spectrum band is exclusively utilized for radio astronomy and geospace sciences. Therefore, RFI detection and mitigation are of critical importance in this mission. 
To identify RFI contamination, SMAP uses multiple statistical RFI detection algorithms. These algorithms are time domain pulse detection, cross-frequency pulse detection, kurtosis detection, and polarization detection, which are applied to both full-band and sub-band data. They all identify RFI based on anomalies relative to the standard deviation of the respective parameter they are scanning for. These algorithms work in conjunction with a logical \textit{'OR'} operation, meaning that a sub-sample is flagged as RFI contaminated if any of these algorithms detect RFI in either the full-band data or the sub-band data \cite{piepmeier2014smap}. Based on the false alarm rate of each of these algorithms, the \textit{'OR'} operation can be deemed as too conservative, leading to a lot of flagged data. Data-driven and DL-based (deep learning-based) approaches have recently been proposed for SMAP's RFI detection by considering the flags as the ground truth for supervised learning \cite{alam2022radio,mohammed2021microwave}. These models can alleviate the need to use multiple detection algorithms. Nevertheless, after detection, in SMAP's RFI mitigation phase, flagged data will be discarded and the average of all data that was not flagged will be calculated. This leads to substantial data loss. To this end, having RFI mitigation methods that preserve valuable information from contaminated data is desired.


The hypothetical case of 5G downlink transmission on the 1.4-1.427 GHz frequency band and its potential impact on SMAP has been recently studied in \cite{9983493} with the goal of identifying the time and amount of interference. While assuming a single or a close cluster of base stations (BSs), this work focuses on recommending the timing schedule for opportunistic access of 5G BSs. However, if this band will no longer be protected with an exclusive license in the future, the RFI from active users will be inevitable, meaning that time scheduling at the active transmitters will not be sufficient to protect remote sensing satellites. Therefore, new reliable and efficient RFI cancellation models are required at SMAP. 

In this paper, we present an autoencoder-based active transmission RFI mitigation method for SMAP's passive radiometer. This method aims to remove the dominant RFI from a known source (e.g., 5G base stations) to potentially preserve the contaminated data from being flagged and discarded by the SMAP's RFI detection and mitigation system. We should note that current DL-based methods on SMAP's RFI detection works are based on real SMAP data where no active transmission is allowed, and the source of RFI is unknown (no ground truth on the RFI source). However, we consider a new approach based on generating ground-truth RFI sources in a simulation environment to have access to both contaminated received signals and dominant RFI signals.

The rest of the paper is as follows. Section \ref{sec:methodology} describes the communication model used in our paper and explains how our autoencoder-based interference mitigation model works. In Section \ref{sec:results}, we describe our simulation experiment and present our results, and Section \ref{sec:conclusion} concludes the paper.


\section{Methodology}
\label{sec:methodology}

\subsection{Communication Model}
Here, we consider a case of coexistence between the SMAP satellite and a terrestrial network, namely, a cluster of 5G base stations in the 1.413GHz frequency. To determine the intensity of RFI caused by the downlink communication of this 5G cluster, we have to identify the attenuation factors of the 5G base station transmissions. As we have unobstructed LoS between SMAP and the 5G cluster, 5G transmissions will be affected by free space path loss and atmospheric attenuation, the latter being negligible compared to the first one \cite{9983493}. The free space path loss is formulated as $ \frac{P_t}{P_r} = G_tG_r (\frac{c}{4\pi f_cd^2})$, 
where $G_t$ is the antenna gain for a 5G base station, $G_r$ is the antenna gain for SMAP's radiometer, and $d$ is the distance between the base station and SMAP. As SMAP operates in orbit with a fixed elevation of 685km from the earth, $d$ depends on the elevation angle.

\begin{figure}[htp]
    \centering
  \includegraphics[clip,width=0.48\columnwidth]{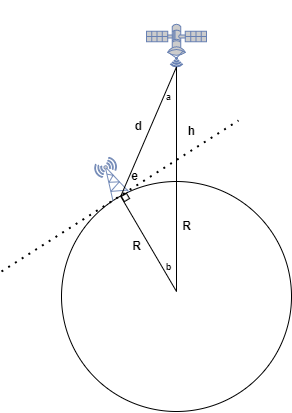}%
\caption{System model. $d$, $e$, $h$, and $R$ represent the distance between the satellite and the base station, elevation angle, satellite elevation, and Earth's radius, respectively.}
\label{fig:satellite_diagram}
\end{figure}

Based on Figure \ref{fig:satellite_diagram}, we can first calculate angle $a$ from the following equation:
\vspace{-10pt}
\begin{align}
\frac{\sin{a}}{R} = \frac{sin{(90+e)}}{R+h}
\end{align}
After that, distance $d$ is calculable as follows:
\vspace{-5pt}
\begin{align}
d = \frac{sin{(90-e-a)}(h+R)}{\sin{(90+e)}}
\end{align}

\vspace{-5pt}

\subsection{Autoencoder-based Interference Mitigation}
\begin{figure}[ht!]
  \includegraphics[clip,width=1\columnwidth]{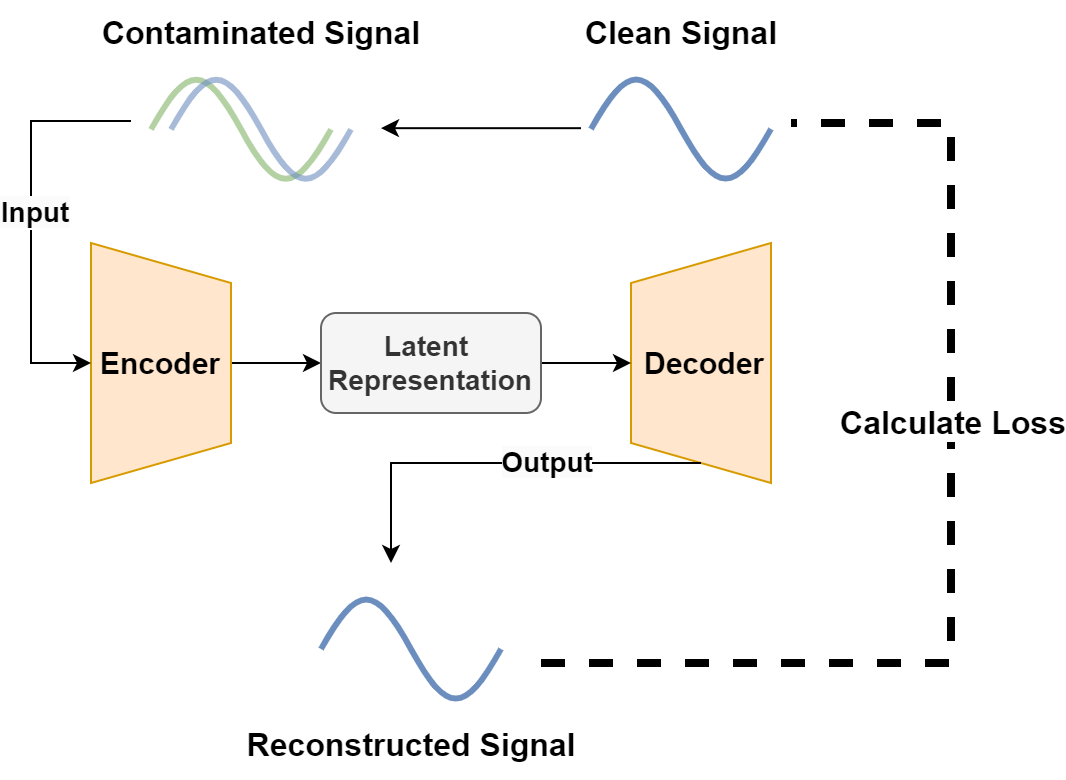}%

\caption{Pipeline of Autoencoder-based interference mitigation.}
\label{fig:denoising_AE}
\end{figure}

\begin{figure}[h!]
  \includegraphics[clip,width=1\columnwidth]{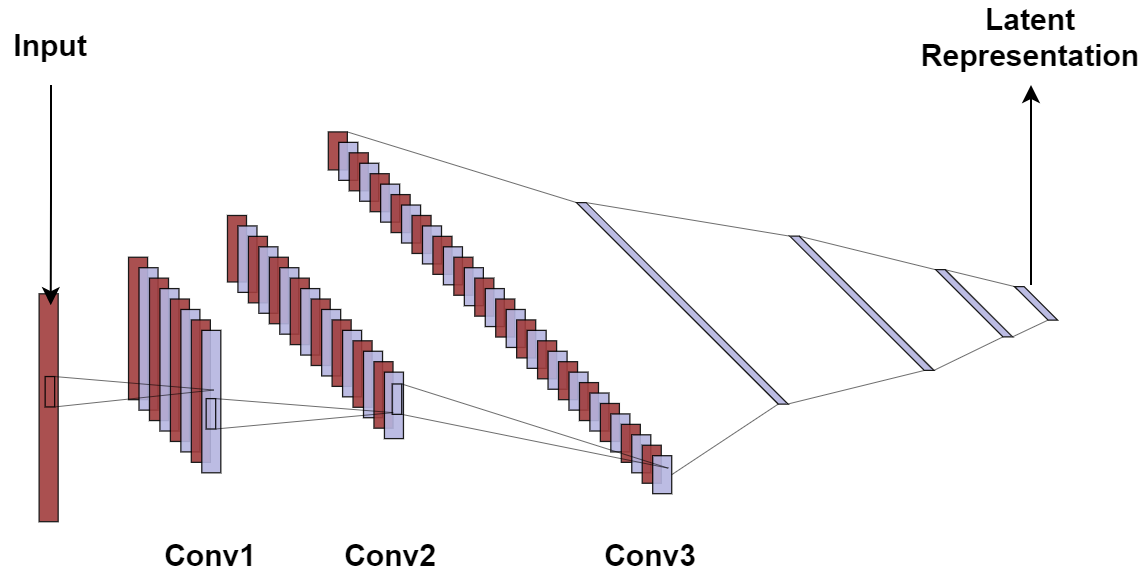}%

\caption{Structure of the encoder. The input has a size of $2\times1000$ where we have 1000 in-phase and 1000 quadrature samples. The latent representation is a 1d vector of size 512.}
\label{fig:encoder_structure}
\end{figure}

Contrary to current SMAP's interference mitigation techniques which focus on detecting and discarding measured signals contaminated by unknown RFI, our AE-based interference mitigation method focuses on reconstructing and deducting the dominant interfering signals of future active users from the received signal at the passive radiometer. Autoencoders have two main components, namely, the encoder and the decoder. The encoder maps the input to a code called the latent representation, which usually has a smaller dimension than the input. The decoder reconstructs the input from the latent representation. In other words, the encoder compresses the input into an encoded representation, from which the input can be reconstructed again. 

Autoencoders have many purposes, but one common utilization of autoencoders is for data denoising, which can be translated to interference mitigation since interference can also be seen as unwanted noise. Fig. \ref{fig:denoising_AE} illustrates the idea in the context of wireless contamination. At first, we have a clean signal which then gets contaminated by another source of wireless transmission. We intend to reconstruct the clean signal from the contaminated signal if possible. The contaminated signal is given as the input to the autoencoder. As opposed to a normal autoencoder, which calculates the loss based on the input and output similarity, the denoising autoencoder compares its output to the initial clean signal. This would push the model to learn how to recreate the clean signal from the contaminated signal.

The structure of the encoder can be seen in Fig. \ref{fig:encoder_structure}. The input is a two-dimensional vector of size $2\times1000$ that represents I/Q (in-phase and quadrature) samples of the contaminated signal (5G plus thermal emission). There are three convolutional layers afterward, all with a kernel size of (1,32). The first has a stride of (1,3), and the next two have a stride of (1,2). After the convolution layers, there are three dense layers with input and output sizes of (3712, 2048), (2048, 1024), and (1024, 512) respective to their order in the structure. Based on the last dense layer, the latent representation has a size of 512. The decoder's structure is basically the reverse of the encoder's structure. There are first three dense layers and then three deconvolutional layers. The decoder takes the latent representation of size 512 as the input and reconstructs the initial I/Q samples with the size of (2,1000).

\section{Simulation and Results}
\label{sec:results}

\begin{figure}[htp]
  \centering
  \includegraphics[clip,width=0.6\columnwidth]{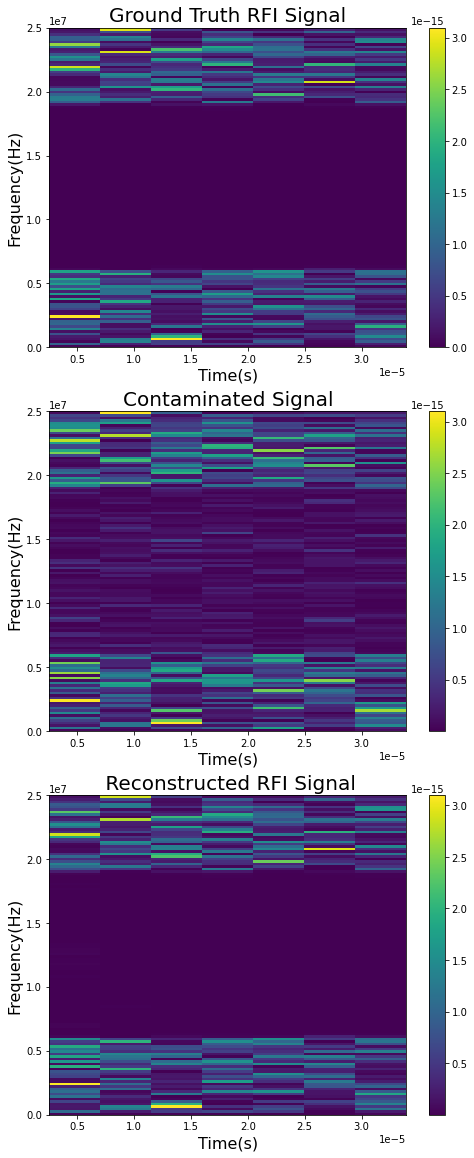}%
\caption{Spectrogram of the ground truth RFI signal (5G), the contaminated signal, and the reconstructed RFI signal (5G) after our autoencoder-based interference mitigation method.}
\label{fig:ae_reconstruction}
\end{figure}

\begin{figure}[h!]
  \includegraphics[clip,width=0.9\columnwidth]{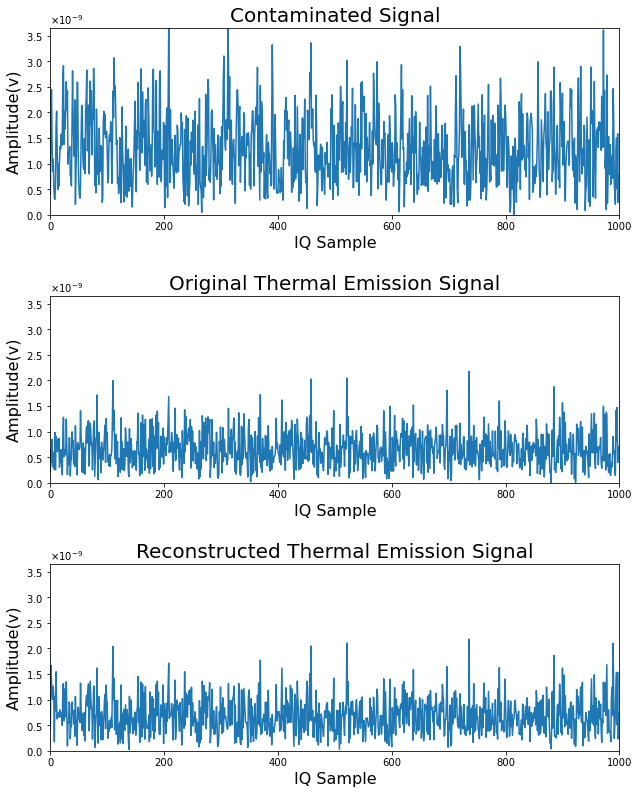}%

\caption{Amplitude of the contaminated signal, the original thermal emission signal, and the reconstructed thermal emission signal.}
\label{fig:amp}
\end{figure}

As mentioned earlier, we use simulated data to generate our dataset. The main reason for this is that we need the contaminated signal as well as the clean signal to train our model. Using the real data collected by SMAP's radiometer, we would have access to whatever the radiometer has sensed, which includes contaminated data in some cases, but we do not have access to the clean data of those instances. SMAP's radiometer includes several measurements, including full-band in-phase and quadrature readings (i.e., I/Q samples); thus, we use MATLAB to generate 5G signal in-phase and quadrature data. Our focus in this paper is to mitigate the RFI at the receiver side (SMAP satellite) and not to optimize the physical layer parameters at the transmitter to minimize the RFI. Thus, we only focused on generating a 5G signal with a reasonable power based on the free space path loss model. With $d=800km$, $f_c=1.4GHz$, and $G_t=12dB$, we calculate the 5G RFI signal's power as -140dB. The thermal emission signal was modeled as a Gaussian signal. Based on the $P = kTB$ formula with $T=300 k$, we calculate the thermal emission's signal power as -144dB.



We are considering severe cases of 5G RFI, where the interfering 5G signal is dominant over the earth's natural thermal emission that SMAP is trying to capture. Fig. \ref{fig:ae_reconstruction} illustrates a spectrogram of the ground truth 5G signal (the interfering signal), the contaminated signal, and the reconstructed 5G signal using our autoencoder-based model. The 5G signal had a power of -50 dB (calculated based on the free space model explained earlier), and the natural emission signal had a power of -54 dB. In the context of SMAP, we care about retrieving the passive radiometer's readings without the dominant interfering signal. To achieve that, we reconstruct the interfering 5G signal from the received contaminated signal and deduct the reconstructed interfering signal from it. This leads to a dominant RFI-free signal which includes the thermal emission readings plus other minor unknown RFI sources, which can be mitigated through SMAP's RFI mitigation techniques.

Figure \ref{fig:amp} compares the amplitude of the reconstructed dominant RFI-free signal against the amplitude of the ground truth of the RFI-free signal and the contaminated signal. As can be observed, the prediction's amplitude is close to the ground truth's amplitude, suggesting that the dominant RFI signal can be mitigated using our autoencoder-based method.

\section{Conclusion}
\label{sec:conclusion}


In this paper, we proposed an autoencoder-based RFI mitigation method for SMAP's passive radiometer. We focused on reconstructing the received signal by extracting the dominant RFI in a scenario where a 5G base station is causing a dominant interference on SMAP's passive radiometer.
We investigated the performance of our proposed RFI mitigation method based on simulation data. Our results suggest that our autoencoder-based interference mitigation method can effectively remove the dominant RFI from the received signal, potentially preserving information that would have otherwise been discarded by the SMAP's interference mitigation method. Our current model can be validated using testbeds that collect remote sensing data with ground truth, as in \cite{farhad2022design}, once they become available.
\bibliographystyle{IEEEbib}
\bibliography{main}

\end{document}